%% file: acl_latex.tex
\documentclass[11pt]{article}
\usepackage{acl}
\usepackage{times}
\usepackage{latexsym}
\usepackage[T1]{fontenc}
\usepackage[utf8]{inputenc}
\usepackage{microtype}
\usepackage{inconsolata}
\usepackage{placeins}
\usepackage{hyperref}
\usepackage{url}
\usepackage{booktabs}
\usepackage{amsmath,amssymb,amsfonts,amsthm}
\usepackage{tikz}
\usetikzlibrary{arrows.meta,positioning,decorations.pathreplacing,calc}
\usepackage{amssymb}
\usepackage[most]{tcolorbox}
\usepackage{multirow}
\usepackage{mathtools} % for \coloneqq and better punctuation
\usepackage{CJKutf8}
\usepackage{relsize} 
\usepackage{tikz}
\usetikzlibrary{arrows.meta,positioning}
\usepackage{subcaption}
\usepackage{array}
\usepackage{algorithm}
\usepackage{algpseudocode}
\usepackage{amsmath}
\usepackage{algorithm}
\usepackage{enumitem}
\usepackage{listings}
\usepackage{xcolor}
\usepackage{graphicx}
\usepackage{tcolorbox}
\usepackage{amsthm}

\tcbuselibrary{listings,skins,breakable}
\title{DGPO: Beyond Pairwise Preferences with Directional Consistent Groupwise Optimization}

\author{
Mengyi Deng$^{1}$, Zhiwei Li$^{1}$, Xin Li$^{1}$, Tingyu Zhu$^{1}$, Yulan Yuan$^{1}$, Zhijiang Guo$^{1,2,\dagger}$, Wei Wang$^{1,2,\dagger}$\\
$^{1}$Information Hub, The Hong Kong University of Science and Technology (Guangzhou), China \\
$^{2}$The Hong Kong University of Science and Technology, Hong Kong SAR\\
\texttt{\{mdeng974, zli404, xli420, tzhu619, yyuan202\}@connect.hkust-gz.edu.cn,} \\
\texttt{zhijiangguo@hkust-gz.edu.cn, weiwcs@ust.hk} \\
}

\begin{document}
\maketitle
\begingroup
\renewcommand\thefootnote{}
\footnotetext{$^\dagger$ Co-corresponding authors.}
\endgroup
\input{sections/0_abstract}
\input{sections/1_intro}

\input{sections/2_relatedwork}

\input{sections/3_method}
\input{sections/4_experiments}

\input{sections/5_conclusion}
\input{sections/6_Limitations}
\bibliography{custom}
\clearpage
\input{sections/7_appendix}

\end{document}

%% file: sections/0_abstract.tex
\begin{abstract}
Although Large Language Models (LLMs) have made remarkable progress, current preference optimization methods still struggle to align directional consistency while preserving reasoning diversity. To address this limitation, we propose \emph{Directional-Groupwise Preference Optimization} (DGPO), a lightweight framework that aggregates supervision signals at the group level and explicitly models direction-aware alignment through multi-candidate comparisons. DGPO organizes forward and reverse question-answer instances into structured sets and optimizes a margin-based likelihood objective that separates coherent reasoning paths from inconsistent alternatives. This groupwise formulation captures richer relative information than pairwise objectives and reinforces consistency across diverse reasoning pathways. Empirical results show that our constructed reverse data yields a 3.2\% average improvement across five benchmarks, while DGPO further delivers consistent gains across multiple datasets and model families, achieving average accuracy improvements of up to 3.6\%. Our code and data are available at \url{https://github.com/Demi-deng2/DGPO}.

\end{abstract}

%% file: sections/1_intro.tex
\section{Introduction}
Large Language Models (LLMs) have demonstrated impressive capabilities across diverse language and reasoning tasks~\citep{openai2023gpt4,touvron2023llama,li2025system}. However, aligning these models to reason faithfully and follow intended problem semantics remains a key challenge. Most existing alignment efforts focus primarily on forward reasoning, where conclusions are derived step by step from given premises—for example, computing the arithmetic mean of all three-digit palindromes (see Figure~\ref{fig:main-framework2}). In contrast, \emph{reverse reasoning}, such as inferring the total sum from the given mean and count of palindromes, has received far less attention. Human cognition is inherently bidirectional: in problem solving, planning, or theorem proving, people routinely combine forward deduction with backward inference~\citep{al2015comparison,newell1972human,jara2019theory}. This motivates the integration of reverse reasoning as a complementary element, essential for achieving more robust and generalizable alignment in LLMs.

Recent work explores reverse supervision as a complement to forward reasoning. MathGenie~\citep{lu2024mathgenie} employs back-translation from solutions to generate new problems, while Reverse Thinking~\citep{chen2024revthink} demonstrates that paired forward and backward exemplars enhance commonsense reasoning. Optimization-focused studies such as OptiBench~\citep{yang2024optibench} extend reverse supervision across linear and nonlinear settings, with ReSocratic applying back-translation of demonstrations. \citet{deng2025inverse} explicitly trains on inverted reasoning trajectories, enabling models to learn bidirectional reasoning patterns rather than relying solely on forward chains.
% For verification and curriculum learning, FOBAR~\citep{jiang-etal-2024-forward} masks problem components to produce backward verification questions, R$^3$~\citep{xi2024r3} shifts start states backward in demonstrations to refine reinforcement signals, and RCOT~\citep{xue2023rcot} reconstructs problems from solutions to expose and correct inconsistencies. 
Broader paradigms further highlight this trend: Reason-from-Future~\citep{xu2025rff} alternates between forward and backward chains to improve math accuracy; backward reconstruction aids causal hypothesis testing~\citep{ranaldi2025backward}; and reverse exemplars strengthen reasoning~\citep{deb2024backward,wang2025sgeu}.

Despite recent advances, existing reverse-supervision methods rarely specify how models should \emph{internalize} directional signals. Reverse exemplars distilled from teacher models are used without explicit directional cues, limiting models’ ability to capture richer alignment signals. This challenge is further compounded by the \emph{Reversal Curse}~\citep{berglund2023reversal}, which shows that transformers often fail to generalize under task inversion due to unstable entity binding, even when trained on large-scale inverted data. Architectural remedies such as the Joint-Embedding Predictive Architecture~\citep{wang2025reversal} mitigate this via additional memory components, but their reliance on strong inductive biases and structural modifications limits scalability. Moreover, most approaches emphasize a single correct solution while overlooking the diversity of valid reasoning paths, reducing models’ ability to generalize across distinct reasoning trajectories. These observations call for alignment strategies that jointly model \emph{directional consistency} and \emph{reasoning diversity} within a unified framework.

We address these challenges with \emph{Directional-Groupwise Preference Optimization (DGPO)}, a framework that builds on a high-quality dataset by generating both forward and reverse supervision, organizing them into structured groups, and optimizing a margin-based likelihood objective. DGPO regulates group preferences through directional consistency, while its groupwise formulation further encourages diversity across reasoning paths. This design effectively distinguishes inconsistent reasoning from coherent alternatives, thereby strengthening the overall alignment signal. Our main contributions are summarized as follows.

\begin{itemize}
\item \textbf{Data Augmentation}: Starting from 817 curated problems in the LIMO dataset~\citep{Ye2025}, we construct groups that contain both forward and reverse instances, with each problem represented by three alternative solution paths in each direction. Distillation on the reverse solutions yields an average accuracy improvement of 3.2\% over the base model.

\item \textbf{Directional-Groupwise Optimization}: We introduce DGPO, which organizes forward and reverse supervision into structured groups and optimizes a margin-based likelihood objective. This groupwise formulation enforces directional consistency while naturally incorporating diverse reasoning paths.

\item \textbf{Empirical Evaluation}: DGPO achieves consistent accuracy gains of 1\%--3.6\% across diverse benchmarks. Additional experiments analyze how scaling the number of reverse groups affects alignment performance.
\end{itemize}

%% file: sections/2_relatedwork.tex
\section{Related Work}
 % Supplementary discussion of data quality and scaling efficiency is provided in Appendix~\ref{app:related-scaling}.
\begin{figure*}
    \centering
    \includegraphics[width=1\linewidth]{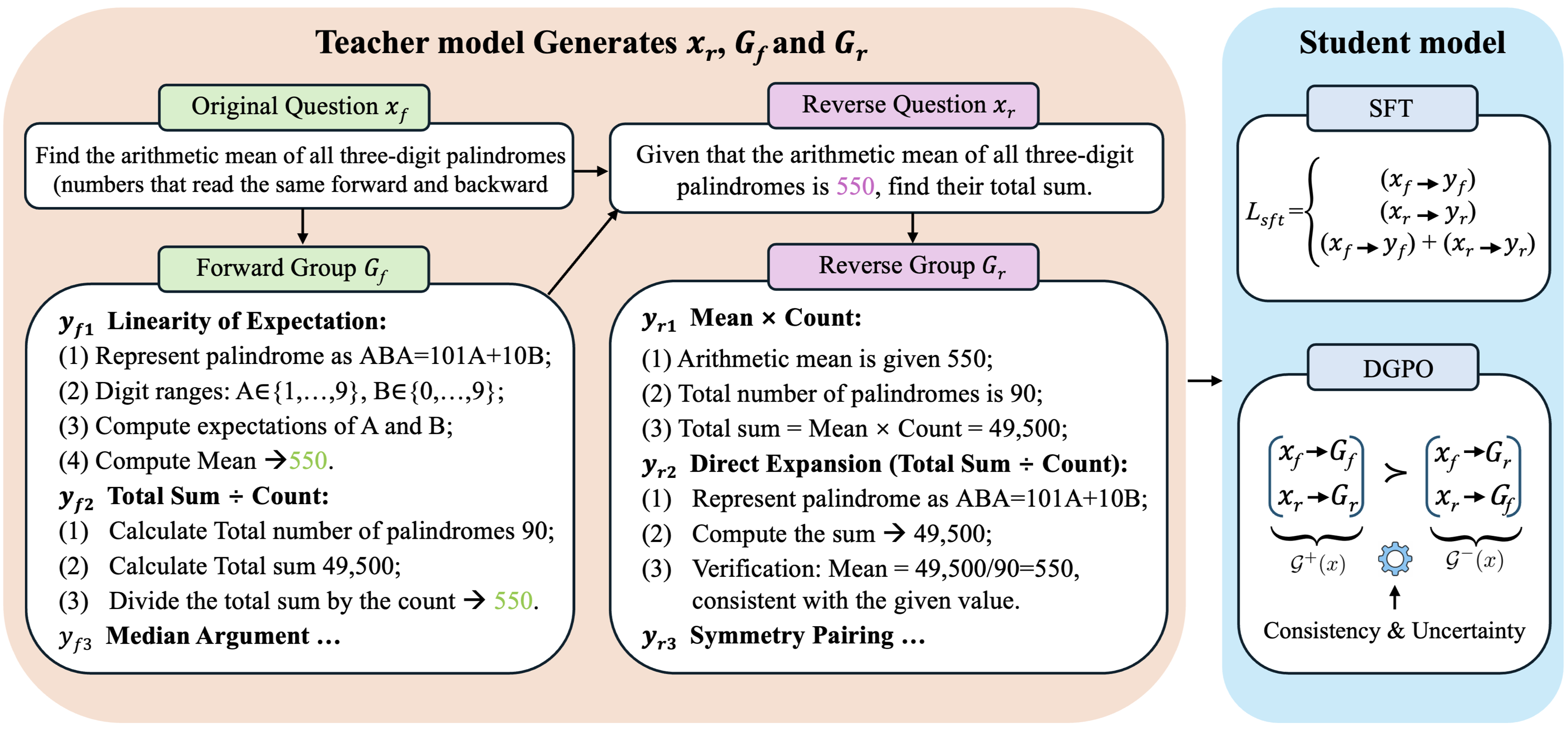}
    \caption{An overview of the DGPO training framework. The process begins with forward problems ($x_f$), each of which can be paired with a reverse question ($x_r$) formulated in the opposite reasoning direction. A teacher model then produces multiple candidate solutions for each problem type ($\{y_{fi}\}_{i=1}^{3}$ for $x_f$ and $\{y_{ri}\}_{i=1}^{3}$ for $x_r$). The solutions are subsequently structured into direction-consistent ($\mathcal{G}^{+}$) and direction-divergent ($\mathcal{G}^{-}$) groups, wherein consistency is determined by matching a prompt's directionality with its corresponding solutions (e.g., $x_f$ with $\{y_{fi}\}_{i=1}^{3}$). DGPO is trained on this structured supervision, incorporating directional modeling and uncertainty-based regulation to enhance alignment stability.}
    \label{fig:main-framework2}
\end{figure*}

\paragraph{Data-Efficient Reverse Supervision.}
Research has increasingly examined reverse reasoning as a complement to forward supervision, highlighting its data efficiency and potential for enhancing model alignment and reasoning diversity (see Appendix~\ref{app:related-scaling} for further discussion). MathGenie~\citep{lu2024mathgenie} generates new problems via solution-to-question back-translation, and RevThink~\citep{chen2024revthink} distills paired forward–backward exemplars. \citet{deng2025inverse} trains on inverted reasoning trajectories, enabling models to learn bidirectional reasoning patterns. Verification-oriented methods such as FOBAR~\citep{jiang-etal-2024-forward} and RCoT~\citep{xue2023rcot} adopt reverse formulations for self-checking, while R$^3$~\citep{xi2024r3} introduces a reverse curriculum in reinforcement learning. OptiBench~\citep{yang2024optibench} and its ReSocratic synthesis strategy construct problems by reversing from solutions. Additional directions include causal hypothesis testing through backward reasoning~\citep{ranaldi2025backward}, reverse-style abductive inference~\citep{deb2024backward}, and reverse exemplars for few-shot prompting and geometry reasoning~\citep{deng2024rcotgeo,wang2025sgeu}. Despite the demonstrated potential of reverse data, its contribution to LLM alignment remains insufficiently studied, particularly in enhancing diversity across reasoning pathways.

\paragraph{Direct Preference Optimization Variants.}
Direct Preference Optimization (DPO)~\citep{Rafailov2023} aligns language models through a closed-form preference objective, avoiding explicit reward-model fitting and unstable online reinforcement learning. Subsequent variants modify either the calibration strategy or the margin structure: KTO~\citep{ethayarajh2024kto} reframes alignment with prospect-theoretic gains and losses, $\beta$-DPO~\citep{wu2024beta} replaces a fixed inverse-temperature with a dynamic $\beta$, and SimPO~\citep{meng2024simpo} removes the reference model through a normalized reward surrogate. Despite these advances, existing DPO variants remain fundamentally pairwise and do not explicitly model whether multiple candidate solutions are directionally coherent with the same problem instance.

\paragraph{Group Relative Policy Optimization}~\citep{shao2024deepseekmath} enhances alignment by operating at the group level, where aggregating multiple outputs provides more diverse preference signals and improves generalization. Several extensions build on this idea: TreeRPO~\citep{yang2025treerpo} improves GRPO by replacing sparse trajectory-level rewards with tree-sampled, step-level dense rewards, enabling more fine-grained optimization of intermediate reasoning steps; Posterior-GRPO~\citep{Fan2025PosteriorGRPO} conditions on successful outcomes to mitigate reward hacking and provide more reliable supervision; and DARS~\citep{yang2025depth} improves GRPO by correcting its bias via difficulty-adaptive rollout sampling and large-batch training to enhance both Pass@K and Pass@1 reasoning performance. Task-oriented refinements include GRPO-LEAD~\citep{Zhang2025GRPOLEAD}, which applies difficulty- and length-aware scaling to encourage concise mathematical reasoning, and noise-aware variants such as S-GRPO~\citep{Shen2025SGRPO}, which use advantage reweighting to improve robustness under imperfect supervision. Despite these advances, applications of group-based optimization to directional alignment remain scarce. This motivates our direction-aware extension, which constructs group-level supervision and leverages internal model judgments to enforce forward-reverse consistency while adaptively modulating update strength based on directional signals.

%% file: sections/3_method.tex
\section{Methodology}\label{sec:method}

We present \emph{Directional-Groupwise Preference Optimization (DGPO)}, a training framework that aggregates supervision signals at the \emph{group} level (see Figure~\ref{fig:main-framework2}). Starting from an original forward problem, the teacher model constructs its reverse counterpart and generates distinct solution paths for both directions, forming a coherent-direction group ($\mathcal{G}^{+}$) and a direction-inconsistent group ($\mathcal{G}^{-}$) (Section~\ref{sec:data}). The student model is then trained under DGPO, where Section~\ref{sec:methohead} introduces directional consistency modeling, and Section~\ref{sec:objective} defines the groupwise training objective.

\subsection{Directional Group Data Construction}
\label{sec:data}

We begin with the 817 curated problems, denoted as forward problems $x_f$, from the LIMO dataset~\citep{Ye2025}, whose small scale and careful curation make it well-suited for studying reasoning alignment. For each $x_f$, the teacher model (DeepSeek V3~\citep{deepseekv3}) generates a corresponding reverse question $x_r$, as illustrated in Figure~\ref{fig:main-framework2}. Subsequently, Qwen3-32B~\citep{qwen3} solves both forward and reverse questions, producing three distinct forward solutions ($\{y_{fi}\}_{i=1}^{3}$) and three reverse solutions ($\{y_{ri}\}_{i=1}^{3}$), each with complete reasoning traces and final answers. To ensure reliability for downstream training, every solution is verified by Qwen3-8B: if an answer is judged incorrect, the corresponding reverse problem is resubmitted to Qwen3-32B until a correct solution is obtained.  

    Based on this pipeline, we organize the data into \emph{directional groups} for DGPO training. For each prompt $x$, we construct a set of preferred, direction-consistent solutions, $\mathcal{G}^{+}(x)$, and a set of dispreferred, direction-divergent solutions, $\mathcal{G}^{-}(x)$. Specifically, for a forward prompt $x_f$, the preferred set is composed of its corresponding forward solutions, $\mathcal{G}^{+}(x_f) = \{y_{fi}\}_{i=1}^{3}$, while the dispreferred set contains the solutions generated for its reverse counterpart, $\mathcal{G}^{-}(x_f) = \{y_{ri}\}_{i=1}^{3}$. Conversely, for a reverse prompt $x_r$, the preferred set consists of its validated reverse solutions, $\mathcal{G}^{+}(x_r) = \{y_{ri}\}_{i=1}^{3}$, and the dispreferred set is formed from the solutions of the original forward problem, $\mathcal{G}^{-}(x_r) = \{y_{fi}\}_{i=1}^{3}$. This bidirectional grouping strategy explicitly encodes the desired reasoning directionality, providing the structured preference signals required for DGPO training. The complete prompt templates and data collection configurations are detailed in Appendix~\ref{app:prompt}.

\subsection{Modeling Directional Consistency}
\label{sec:methohead}

Having constructed bidirectional groups of forward and reverse solutions, we train the model to discern directionally coherent reasoning from inconsistent alternatives. Each prompt includes multiple candidate responses, comprising direction-consistent solutions and reverse-problem solutions that appear off-target within the current reasoning context. Since these candidates reflect varying degrees of directional consistency, we introduce a confidence-aware mechanism that estimates directional alignment together with an associated uncertainty signal.

\subsubsection{Consistency Estimation}
\label{sec:posterior}

For each prompt-solution pair \((x, y)\), we process the concatenated sequence \((x; y)\) using the policy model \(\pi_{\theta}\) to obtain a hidden representation \(h_{\theta}(x,y)\). This representation is fed into a small projection head that predicts the parameters \((\alpha, \beta)\) of a Beta distribution. This distribution models the uncertainty in our estimate of the probability that solution \(y\) is directionally consistent with prompt \(x\). We denote the resulting Beta-parameterized consistency distribution as:
\[
q(d \mid x, y) = \mathrm{Beta}(\alpha(x, y), \beta(x, y)),
\]
where \(\alpha(x, y) > 0\) and \(\beta(x, y) > 0\) are the positive outputs of the projection head. The mean of this distribution, \(d(x, y) = \mathbb{E}[d] = \alpha/(\alpha + \beta)\), represents the estimated probability that the solution \(y\) is directionally consistent with the prompt \(x\). The variance, \(\sigma^2(x, y) = \mathrm{Var}[d] = \alpha\beta / [(\alpha+\beta)^2(\alpha+\beta+1)]\), quantifies the uncertainty proxy in this estimate. To stabilize training, we prevent gradients from flowing from the consistency head back to the policy encoder's hidden states \(h_\theta(x, y)\) using a stop-gradient operation.

\subsubsection{Variance-Aware Aggregation}
\label{sec:weighting}

The estimated consistency and uncertainty are used to compute a weighted influence for each solution during group aggregation. While the framework theoretically supports multiple formulations for combining these signals, our implementation adopts a numerically stable variant where variance penalties are directly incorporated into the preference scores rather than the weights.

For each solution, we define the preference score as the length-normalized log-probability under the policy:
\begin{equation}
\label{eq:delta}
% \Delta_{\theta}(y \mid x) \;=\; \frac{1}{|y|}\,\log \pi_{\theta}(y \mid x),
\Delta_{\theta}(y \mid x) \;=\; \frac{1}{|y|} \sum_{t=1}^{|y|} \log \pi_{\theta}\!\left(y_t \mid x, y_{<t}\right),
\end{equation}
where $|y|$ denotes the number of tokens in the response $y$. We compute the log-probability over response tokens only. This length normalization prevents systematic bias towards longer responses.

The pre-activation scores for the preferred and dispreferred groups are computed by combining the policy's likelihood, the estimated consistency, and an uncertainty penalty:
\begin{equation}
\label{eq:preact}
\scalebox{0.82}{%
$\begin{aligned}
u^{+}(x, y) &= \tau_{\mathrm{win}}^{-1}\Delta_{\theta}(y \mid x) + \log d(x, y) -  \sigma^{2}(x, y), \\
u^{-}(x, y) &= \tau_{\mathrm{lose}}^{-1}\Delta_{\theta}(y \mid x) + \log (1 - d(x, y)) - \sigma^{2}(x, y),
\end{aligned}$%
}
\end{equation}

where $\tau_{\mathrm{win}}, \tau_{\mathrm{lose}} > 0$ are temperature parameters. This formulation rewards solutions with high likelihood ($\Delta_{\theta}$), high predicted directional consistency ($\log d(x, y)$ or $\log(1-d(x, y))$), and low estimated uncertainty (as directly penalized by the variance term $-\sigma^{2}(x, y)$).

\subsection{DGPO Training Objective}
\label{sec:objective}

The pre-activation scores for all solutions in a group are aggregated to form group-level scores using a temperature-scaled log-sum-exp operation:
\begin{equation}
\label{eq:aplus}
A_{\theta}^{+}(x) = \tau_{\mathrm{win}} \log \sum_{y \in \mathcal{G}^{+}(x)} \exp\left(u^{+}(x, y)\right),
\end{equation}
\begin{equation}
\label{eq:aminus}
A_{\theta}^{-}(x) = \tau_{\mathrm{lose}} \log \sum_{y \in \mathcal{G}^{-}(x)} \exp\left(u^{-}(x, y)\right).
\end{equation}

The core training objective is a contrastive loss that encourages a margin between the aggregated score of the preferred group and that of the dispreferred group:
\begin{equation}
\label{eq:dgpo}
\scalebox{0.86}{%
$\begin{aligned}
\mathcal{L}_{\mathrm{DGPO}}(\theta, \phi) = -\mathbb{E} \left[ \log \sigma\left(  \lambda_{\mathrm{margin}} \left( A_{\theta}^{+}(x) - A_{\theta}^{-}(x) \right) \right) \right],
\end{aligned}$%
}
\end{equation}
where $\lambda_{\mathrm{margin}} > 0$ is a scaling factor and $\sigma$ is the logistic sigmoid function.

To mitigate model overconfidence and stabilize the training of the consistency head, we introduce two regularization terms. The first is a directional KL divergence penalty that incorporates prior knowledge of group structure by discouraging deviations from group-specific prior distributions:

\begin{equation}
\label{eq:rkl}
\begin{aligned}
\mathcal{R}_{\mathrm{KL}} = \lambda_{\mathrm{kl}} \mathbb{E}_{x, y} \big[
    &\mathbb{I}(y \in \mathcal{G}^{+}) \cdot \mathrm{KL}\left( q \parallel p_{+} \right) \\
    + &\mathbb{I}(y \in \mathcal{G}^{-}) \cdot \mathrm{KL}\left( q \parallel p_{-} \right) \big],
\end{aligned}
\end{equation}

where $q = q(\cdot \mid x, y)$ denotes the learned consistency distribution, $\mathbb{I}(\cdot)$ is the indicator function, and $p_{+}$ and $p_{-}$ are asymmetric Beta priors corresponding to the preferred and dispreferred groups, respectively. These priors are chosen such that $p_{+}$ is biased toward 1, encouraging high consistency estimates for $\mathcal{G}^{+}$, while $p_{-}$ is biased toward 0, favoring low consistency estimates for $\mathcal{G}^{-}$.

The second is an optional penalty on the average predictive uncertainty across all candidate solutions for a prompt:
\begin{equation}
\label{eq:rvar}
\mathcal{R}_{\mathrm{var}} = \lambda_{\mathrm{var}}^{\mathrm{(grp)}} \mathbb{E}_{x} \left[ \frac{1}{|\mathcal{G}(x)|} \sum_{y \in \mathcal{G}(x)} \sigma^{2}(x, y) \right],
\end{equation}
where $\mathcal{G}(x) \triangleq \mathcal{G}^{+}(x) \cup \mathcal{G}^{-}(x)$ denotes the full set of candidate solutions for prompt $x$.

The complete objective function minimized during training is:
\begin{equation}
\label{eq:final}
\mathcal{J}(\theta, \phi) = \mathcal{L}_{\mathrm{DGPO}}(\theta, \phi) + \mathcal{R}_{\mathrm{KL}} + \mathcal{R}_{\mathrm{var}}.
\end{equation}

The parameters of the policy model $\theta$ and the consistency head $\phi$ are optimized jointly to minimize the objective $\mathcal{J}$. We employ the AdamW optimizer with standard practices, including gradient clipping and a cosine learning rate decay schedule. The base training procedure is summarized in Algorithm~\ref{alg:dgpo}. The algorithm processes minibatches of prompts, and for each prompt, it computes the group aggregates and updates the model parameters. For specific implementation details, including hyperparameters and model configurations, we refer the reader to Section~\ref{head}.

\begin{algorithm}[t]
\caption{DGPO Training}
\label{alg:dgpo}
\begin{algorithmic}[1]
    \State \textbf{Input:} Minibatch $\mathcal{B}$, policy $\pi_{\theta}$, consistency head $q_{\phi}$, hyperparameters $\tau_{\mathrm{win}},\tau_{\mathrm{lose}},\lambda_{\mathrm{kl}},\lambda_{\mathrm{var}}^{\mathrm{(grp)}}$

    \State \textbf{Output:} Updated parameters $\theta,\phi$

    \For{each prompt $x \in \mathcal{B}$}
        \State Retrieve $\mathcal{G}^{+}(x)$ and $\mathcal{G}^{-}(x)$
        \For{each $y \in \mathcal{G}^{+}(x) \cup \mathcal{G}^{-}(x)$}
            \State Compute $h_{\theta}(x,y)$ 
            \State Obtain $d(x,y)$, $\sigma^{2}(x,y)$ from $q_{\phi}$
            \State Compute $\Delta_{\theta}(y\mid x)$ via Eq.~\eqref{eq:delta}
            \State Form $u^{+}(x,y), u^{-}(x,y)$ via Eq.~\eqref{eq:preact}
        \EndFor
        \State Aggregate $A_{\theta}^{+}(x), A_{\theta}^{-}(x)$ via Eqs.~\eqref{eq:aplus}--\eqref{eq:aminus}
        \State Compute $\mathcal{L}_{\mathrm{DGPO}}(x)$ via Eq.~\eqref{eq:dgpo}
        \State Compute $\mathcal{R}_{\mathrm{KL}}(x)$ via Eq.~\eqref{eq:rkl}
        \State Compute $\mathcal{R}_{\mathrm{var}}(x)$ via Eq.~\eqref{eq:rvar}
    \EndFor
    \State $\mathcal{J} \gets  \frac{1}{|\mathcal{B}|}\sum_{x\in\mathcal{B}}[\mathcal{L}_{\mathrm{DGPO}}(x) + \mathcal{R}_{\mathrm{KL}}(x) + \mathcal{R}_{\mathrm{var}}(x)]$
    \State Update $\theta \gets \theta - \eta \nabla_{\theta} \mathcal{J}$
    \State Update $\phi \gets \phi - \eta \nabla_{\phi} \mathcal{J}$
    \State \Return $\theta,\phi$
\end{algorithmic}
\end{algorithm}

\label{sec:theory}

When each group $\mathcal{G}^{+}(x)$ and $\mathcal{G}^{-}(x)$ contains only one response, Eqs.~(\ref{eq:aplus})--(\ref{eq:dgpo}) collapse to the familiar pairwise contrastive objective, showing that DGPO can be viewed as a groupwise extension of conventional preference optimization. When multiple responses are available, the log-sum-exp aggregation in Eqs.~(\ref{eq:aplus}) and (\ref{eq:aminus}) acts as a smooth surrogate for selecting the higher-scoring responses from each group, allowing several candidates to contribute without introducing the discontinuity of a hard maximum. Finally, the Beta-parameterized consistency head provides a confidence-aware mechanism for modulating each response before group aggregation, so DGPO favors responses that are not only likely under the policy but also more aligned with the intended reasoning direction.

% The DGPO objective admits a simple probabilistic interpretation that highlights its stability and consistency. 
% First, when each group $\mathcal{G}^{+}(x)$ and $\mathcal{G}^{-}(x)$ contains only a single solution, 
% Eqs.~(\ref{eq:aplus})--(\ref{eq:dgpo}) reduce to a standard pairwise contrastive form, 
% ensuring compatibility with conventional preference optimization. 
% Second, the logarithmic soft-max aggregation in Eqs.~(\ref{eq:aplus}) and (\ref{eq:aminus}) provides a differentiable approximation to the maximum groupwise margin, ensuring smooth and stable optimization. 
% Finally, by interpreting the posterior $q(d\!\mid\!x,y)$ as the distribution over directional consistency, 
% the objective in Eq.~(\ref{eq:final}) can be viewed as maximizing the expected likelihood of alignment under epistemic uncertainty. 
% Formal propositions and proofs are provided in Appendix~\ref{app:theory}.

%% file: sections/4_experiments.tex
\section{Experiments}

\subsection{Experimental Setup}
\label{exp-setup}
\paragraph{Training Configuration.}
All experiments are implemented using the \texttt{SWIFT} framework~\citep{zhao2024swiftascalablelightweightinfrastructure}. We first perform supervised fine-tuning (SFT) for 3 epochs with a learning rate of $1\times10^{-5}$ and a maximum sequence length of 11,000 tokens. Subsequently, DGPO training is conducted for 2 epochs with learning rates ranging from $1\times10^{-6}$ to $3\times10^{-6}$ and a maximum sequence length of 1,000 tokens. All models are trained in \textbf{bfloat16} precision on four NVIDIA A6000 GPUs (48GB each). Additional training details are provided in Appendix~\ref{app:impl}.

\begin{table*}[h]
\centering
\caption{Performance comparison of DGPO across two model families: Qwen3-1.7B-Base and SFT models trained on mixed forward–reverse data followed by DGPO. 
We also compare different SFT dataset configurations, where 1Mixed denotes equal-size blending of forward (LIMO) and reverse data, and 3Mixed triples the reverse portion. None denotes the official pretrained model without any additional fine-tuning or preference optimization. The best result is highlighted in \textbf{bold}, and the second best is \underline{underlined}.}
\label{tab:main-results}
\resizebox{\textwidth}{!}{
\begin{tabular}{llrrrrrr}
\toprule
\textbf{Base Model} & \textbf{Training Strategy}
& \textbf{AIME-25} 
& \textbf{GPQA} 
& \textbf{Math 500} 
& \textbf{GMQ} 
& \textbf{LMGH} 
& \textbf{Avg. Acc.} \\
\midrule
Qwen3-1.7B-Base   & None      & 0\%     & 28.3\%   & 45.8\%   & 33.6\% & 2.3\% & 22\%\\
Qwen3-1.7B-Base   & SFT (LIMO)      & 0\%   & 27.8\%    & 47.2\% & 34.8\%  &  \underline{11.4}\% & 24.2\%\\
Qwen3-1.7B-Base   & SFT (Reverse)   &3.3\%  & \underline{29.8}\% & 46.2\%    & 33.6\%  &  \textbf{12.9\%} &  \underline{25.2\%} \\
Qwen3-1.7B-Base   & SFT (1Mixed)     & 0\%     & 29.3\% & \underline{46.2\%}   & \textbf{35.3\%}  &9.9\% &24.1\%\\
Qwen3-1.7B-Base   & SFT (3Mixed)     & 0\%     & 27.3\% & 43.2\%   & 33.9\%  &11.3\% & 23.1\%\\
SFT (3Mixed) & Vanilla DPO            & 0\%     & 27.8\%  & 39.2\%   & 30.3\% &4.5\% &20.4\% \\
SFT (3Mixed) & DGPO (Ours)  & \underline{6.7\%}  & \textbf{30.3\%} & 43.6\% & 33.6\% & 12.1\% &\textbf{25.3\%} \\
% \midrule
% Qwen3-1.7B-Base & None     & 0\%     & 28.3\%   & 45.8\%   & 33.6\% & 2.3\% & 22\%\\
Qwen3-1.7B-Base & DGPO (Ours)   & \textbf{10}\%  & 28.8\% & \textbf{46.6\%} & \underline{35\%}  & 3.0\% & 24.7\%\\
% \midrule
% Qwen3-1.7B  & None   &13.3\%  & 29.8\%  &  47.6\%  &  36.2\% &  10.6\% & 27.5\%\\
% Qwen3-1.7B & DGPO (Ours)  & 26.7\% & 32.3\%  &  46\%  & 36.5\%  & 12.9\%  & 30.9\%  \\
\bottomrule
\end{tabular}
}
\end{table*}

\paragraph{Evaluation.}
We employ five distinct benchmarks to thoroughly assess the reasoning alignment performance, covering both in-distribution performance and generalization capabilities. The OpenAI Math 500 dataset~\citep{Lightman2023LetsVerify,hendrycks2021measuring} is adopted as an \emph{in-domain} benchmark, as it closely aligns with the algebraic, geometric, and symbolic reasoning characteristics present in LIMO. AIME-25~\citep{aime_dataset} is considered a \emph{near-domain} benchmark, sharing a competition-style format with LIMO while exhibiting differences in problem distribution and year-specific variations. To further evaluate generalization, we incorporate three \emph{out-of-domain} benchmarks: GPQA~\citep{clark2023gpqa}, which assesses graduate-level scientific and conceptual reasoning; Gaokao MathQA (GMQ)~\citep{gaokao_mathqa}, comprising high-school level mathematical word problems; and Leaderboard Math: Math Geometry Hard (LMGH)~\citep{leaderboard_math}, a curated set of challenging geometry problems. Collectively, these benchmarks encompass diverse domains, difficulty levels, and problem types, providing a comprehensive evaluation framework for reasoning alignment.

Our evaluation reports the pass@1 accuracy in a zero-shot chain-of-thought setting. We generate reasoning paths using greedy sampling, setting the maximum output length to 16,000 tokens for AIME-25 and 10,000 tokens for the other benchmarks to prevent truncation. The evaluation process follows the standard lm-eval-harness~\citep{lm-eval-harness}, leveraging its rule-based matching system that incorporates canonicalized string normalization and numerical equivalence verification. This consistent protocol guarantees an objective and fully reproducible evaluation across the diverse set of reasoning benchmarks.
\subsection{Effectiveness of Group Data and DGPO}

\begin{table*}[h]
\centering
\caption{Performance comparison of DGPO and other DPO variants on Qwen3-1.7B-Base. The best result in each metric is highlighted in \textbf{bold}.}
\label{tab:dpo-variants}
\begin{tabular}{llrrrrrr}
\toprule
\textbf{Base Model} & \textbf{DPO Variant} & \textbf{AIME-25} & \textbf{GPQA} & \textbf{Math 500} & \textbf{GMQ} & \textbf{LMGH} & \textbf{Avg. Acc.} \\
\midrule
Qwen3-1.7B-Base & $\beta$-DPO & 0\% & 27.3\% & 47.6\% & 35.6\% & 3.8\% & 22.9\%\\
Qwen3-1.7B-Base & $\gamma$-DPO & 0\% & 28.3\% & 45.6\% & 35.0\% & 1.5\% & 22.1\%\\
Qwen3-1.7B-Base & SimPO & 0\% & 27.8\% & 47.0\% & 33.3\% & 3.0\% & 22.2\%\\
Qwen3-1.7B-Base & Ours & \textbf{10.0\%} & \textbf{28.8\%} & 46.6\% & 35.0\% & 3.0\% & \textbf{24.7\%}\\
\bottomrule
\end{tabular}
\end{table*}

\begin{table*}[h]
\centering
\caption{
Ablation study on different training settings. 
\textbf{Ours} employs variance regularization $\mathcal{R}_{\mathrm{var}}$ and a differentiable posterior head. 
We ablate these components by removing the variance regularization (w/o $\mathcal{R}_{\mathrm{var}}$) 
and by disabling the differentiable posterior (w/o Posterior). 
Percentages with $\downarrow$ indicate the average performance drop relative to \textbf{Ours}. None denotes the official model without further fine-tuning or alignment training. The best results within each base model family are highlighted in \textbf{bold}.}
\label{tab:ablation study}
\resizebox{\textwidth}{!}{
\begin{tabular}{llrrrrrr}
\toprule
\textbf{Base Model} 
& \textbf{Setting} 
& \textbf{AIME-25} 
& \textbf{GPQA} 
& \textbf{Math 500} 
& \textbf{GMQ} 
& \textbf{LMGH} 
& \textbf{Avg. Acc.} \\
\midrule
\multirow{4}{*}{Qwen3-1.7B-Base} & None      & 0\%     & 28.3\%   & 45.8\%   & 33.6\% & 2.3\% & 22\%\\
&  Ours & \textbf{10\%}  & \textbf{28.8\%} & 46.6\% & 35\%  & 3.0\% & \textbf{24.7\%}\\
 &  ~~~~~w/o $\mathcal{R}_{\mathrm{var}}$    & 3.3\% &   \textbf{28.8\%} &  \textbf{46.8\%}  & \textbf{32.5}\%  & 3.0\%  & 22.9\%(1.8\%$\downarrow$) \\
 & ~~~~~w/o Posterior &   0\%     & 27.8\%   & 45\%   & 33.0\% & 3.0\% & 21.8\%(2.9\%$\downarrow$)\\
% \midrule
% \multirow{4}{*}{Qwen3-1.7B} & 
%  None   &13.3\%  & 29.8\%  &  \textbf{47.6\%}  &  36.2\% &  10.6\% & 27.5\%\\
% &Ours & \textbf{26.7\%} & \textbf{32.3\%}  &  46\%  & 36.5\%  & 12.9\%  & \textbf{30.9\%}  \\
%  & ~~~~~w/o $\mathcal{R}_{\mathrm{var}}$   & 10\% & 31.3\%  & 46.2\% & 35.9\% &   12.9\% &  27.3\%(3.6\%$\downarrow$) \\
%  & ~~~~~w/o Posterior  & 16.7\% & 30.8\%  &  44.6\%  &  \textbf{37\%} & 12.9\%  & 28.3\%(2.6\%$\downarrow$)  \\
\bottomrule
\end{tabular}
}
\end{table*}

Before evaluating the effectiveness of DGPO, we first examine the quality and behavioral impact of the constructed reverse data through supervised fine-tuning (SFT). 
Experiments are conducted mainly on Qwen3-1.7B-Base. 
We distill models on the forward (LIMO) and reverse datasets independently, as well as on their mixed combinations, except for the 3Mixed setting, each containing problems with exactly one verified reasoning path. As shown in Table~\ref{tab:main-results} and Appendix~\ref{app:reverse-quality}, the reverse subset data demonstrates comparable quality to the original forward LIMO data. For the Qwen3-1.7B-Base, SFT with reverse data achieves an average accuracy of 25.2\%, slightly higher than forward-only SFT (24.2\%), confirming that the reverse counterparts provide meaningful and learnable supervision. 

However, directly mixing forward and reverse data results in weaker performance compared to single-direction SFT. The Mixed setting drops to 24.1\%, and increasing the ratio of reverse data in the 3Mixed setting further reduces performance to 23.1\%.  This suggests that naive data blending introduces interference between opposite reasoning directions rather than synergy. A similar pattern appears for the Qwen3-1.7B model (see Appendix~\ref{app:reverse-quality}), where distillation on reverse subsets or mixed data causes noticeable drops across benchmarks. While the constructed groupwise data derived from forward and reverse subsets is of high quality, these results reveal that naive joint training across opposite directions introduces conflicts. This shows the need for an alignment methodology that can effectively leverage groupwise supervision while preserving the diversity of directional reasoning.

We evaluate DGPO on three model foundations: Qwen3-1.7B-Base, SFT model trained on the 3Mixed subsets, and Qwen3-1.7B combining forward and reverse data. 
DGPO is trained on 1,634 groupwise preference instances, where each forward and reverse counterpart form directional contrasted groups, enforcing consistency between preferred and dispreferred reasoning directions. As shown in Table~\ref{tab:main-results}, DGPO delivers consistent performance improvements across all model families.
On Qwen3-1.7B-Base, DGPO improves the average accuracy from 22.0\% to 24.7\%, confirming that groupwise optimization provides alignment benefits beyond direct fine-tuning.
When applied to the mixed-direction SFT model (3Mixed), DGPO further strengthens results by  2.2\% and clearly outperforms vanilla DPO, demonstrating its effectiveness in mitigating inconsistencies introduced by multi-directional training. 

For Qwen3-1.7B reported in  Table~\ref{tab:Direction impact}, DGPO improves the average accuracy from 27.5\% to 30.9\%, with substantial gains on AIME-25 and GPQA, indicating stronger generalization to near and out-of-distribution tasks. However, a slight decline on Math 500 suggests a modest trade-off for in-distribution performance as the model learns to balance broader reasoning consistency. These results demonstrate that DGPO effectively integrates groupwise supervision across reasoning directions and strengthens alignment even on highly optimized models. To further validate the robustness and reliability of our results, we repeat each DGPO experiment three times with different random seeds and report the mean accuracy and standard deviation across runs in Table~\ref{tab:multi-run-study}.

\begin{table*}[h]
\centering
\caption{Effect of the number of reverse problem groups on model performance. 
\textbf{Number} indicates the group configuration: 0 denotes the pretrained model without additional training, 2 corresponds to the LIMO forward set paired with one reverse group, 3 with two, and 4 with three. Within each base model family, the best result is shown in \textbf{bold} and the second-best is \underline{underlined}. 
An upward arrow (\(\uparrow\)) indicates the relative improvement in average accuracy over the base model.}
\label{tab:Direction impact}
\resizebox{\textwidth}{!}{
\begin{tabular}{lcrrrrrr}
\toprule
\textbf{Base Model} & \textbf{Number}  
& \textbf{AIME-25} 
& \textbf{GPQA} 
& \textbf{Math 500} 
& \textbf{GMQ} 
& \textbf{LMGH} 
& \textbf{Avg. Acc.} \\
\midrule

\multirow{4}{*}{Qwen3-1.7B-Base}  & 0     & 0\%     & 28.3\%   & 45.8\%   & 33.6\% & 2.3\% & 22\%\\
% Qwen3-1.7B  & Official model   &13.33\%  & 29.79\%  &  46.2\%  &  35.9\% &  10.61\% & 27.17\%\\
% \midrule
 & 2  & \textbf{10}\%  & 28.8\% & 46.6\% & \underline{35\%}  & 3.0\% & \textbf{24.7\%}(2.7\%$\uparrow$)\\
 & 3 &  3.3\%   & \textbf{29.8}\%  &  \underline{47.0}\%  & 33.9\%  & \underline{3.8\%}  &  23.6\%(1.6\%$\uparrow$)  \\
 & 4 & \underline{6.7\%}   & \underline{29.3}\% & \textbf{47.8\%}  &  \textbf{35.6}\% & \textbf{4.5\%}  &  \underline{23.9\%}(1.9\%$\uparrow$)  \\
\midrule
\multirow{4}{*}{Qwen3-1.7B}  & 0    &13.3\%  & 29.8\%  &  \underline{47.6\%}  &  36.2\% &  10.6\% & 27.5\%\\
 &2   &  \textbf{26.7\%} & 32.3\%  &  46\%  & \underline{36.5\%}  & 12.9\%  & \underline{30.9\%} (3.4\%$\uparrow$) \\
 &3   & 20.0\%  &  \textbf{34.8\%}  &  \textbf{48.0\%} &36.2\% & \textbf{16.7\%} &  \textbf{31.1\%}(3.6\%$\uparrow$)  \\
 &4  & \underline{23.3}\% & \underline{32.8\%}  &  46.6\%  & \textbf{37}\% &  \underline{13.6\%}  &  30.7\%(3.2\%$\uparrow$) \\
\bottomrule
\end{tabular}
}
\end{table*}

As shown in Table~\ref{tab:dpo-variants}, DGPO achieves the strongest average accuracy on Qwen3-1.7B-Base. Compared with $\beta$-DPO~\citep{wu2024beta} and $\gamma$-DPO~\citep{sun2025robust}, DGPO consistently improves AIME-25 while remaining competitive on GPQA and GMQ, indicating that modeling directional consistency at the group level yields a more reliable alignment signal than dynamically rescaling pairwise margins alone. Compared with SimPO~\citep{meng2024simpo}, DGPO also improves out-of-domain generalization on GPQA and maintains a stronger balance across benchmarks, suggesting that removing the reference model is insufficient without explicitly separating direction-consistent from direction-divergent reasoning paths.

Table~\ref{tab:ablation study} presents a detailed ablation analysis of the core components in DGPO. The base configuration (Ours) integrates both the variance regularization term $\mathcal{R}_{\mathrm{var}}$ and a differentiable posterior head that explicitly models directional consistency within each group. When the posterior head is removed, the training reduces to an offline GRPO-style objective, 
where preference scores within each group are computed solely from the averaged log probabilities $\log \pi_{\theta}(y|x)$ and combined through a smooth maximum operation, without explicitly modeling directional alignment or uncertainty. Even though this simplified variant still brings moderate improvements, the overall gains are clearly smaller than those achieved by the full DGPO framework, with the most evident declines observed on AIME-25. 

In contrast, removing the variance regularization term $\mathcal{R}_{\mathrm{var}}$ 
leads to a moderate performance decrease of 1.8\%, while removing the posterior head results in reductions of 2.9\%. These findings suggest that penalizing predictive uncertainty helps stabilize groupwise optimization and that directional consistency modeling provides informative signals for distinguishing coherent from divergent reasoning patterns. Both uncertainty regularization and explicit directional modeling are key components contributing to DGPO’s stable and robust improvements over standard groupwise contrastive training.

\subsection{Scaling Effects of Reverse Group Augmentation}

To examine how the quantity of reverse problem groups influences alignment, we systematically vary the number of reverse sets paired with each forward exemplar. 
Each forward problem in the LIMO dataset can be paired with multiple reverse questions generated by DeepSeek V3, where each problem (forward or reverse) is accompanied by three native solutions produced by Qwen3-32B (see Appendix~\ref{app:prompt} for data construction details).
For DGPO training, the preferred group $\mathcal{G}^{+}(x)$ for a given problem $x$ consists of its direction-consistent solutions, while for the scaling experiment, the dispreferred group $\mathcal{G}^{-}(x)$ is formed by selecting direction-divergent solutions drawn from unrelated or mismatched problem instances. 
This design enables a controlled investigation of group-level augmentation while maintaining consistent supervision quality.

As shown in Table~\ref{tab:Direction impact}, the effect of increasing reverse groups exhibits distinct trends across model families. 
For Qwen3-1.7B-Base, moderate augmentation enhances alignment, but adding more groups leads to progressively smaller gains. Performance peaks when introducing a single reverse group, suggesting that mild directional diversification most effectively strengthens the model’s reasoning alignment. However, further augmentation leads to a slight decline, implying that excessive directional mixing introduces representational interference and hinders stable generalization. This consistent decline in performance indicates that low-capacity base models struggle to accommodate increasingly diverse directional signals.

In contrast, the Qwen3-1.7B model exhibits a more stable and scalable improvement trend. Performance increases steadily with additional reverse groups, reaching a peak of 31.1\% when two reverse groups are included. Most of the gains come from GPQA, Math 500, and LMGH, implying that expanding reverse-group data particularly strengthens reasoning generalization. When the number of reverse groups increases further, only modest gains are observed on GMQ and LMGH, while the overall performance falls slightly below that of the two-group configuration, indicating that the benefits of additional augmentation are limited. Taken together, these results indicate that while expanding the number of reverse groups yields measurable improvements in reasoning alignment, the gains gradually saturate rather than continue to scale with data size.

%% file: sections/5_conclusion.tex
\section{Conclusion}

This paper introduces DGPO, a framework that models directional consistency while preserving intra-group diversity. Built on curated forward–reverse exemplars, it constructs structured groups capturing coherent and contrastive reasoning behaviors for finer alignment. Empirical results demonstrate that DGPO consistently improves alignment across multiple model families, achieving accuracy gains of 2.2\%–3.6\%. 
Ablation studies confirm that both uncertainty regularization and direction-aware modeling are beneficial, as removing either leads to a 1.8\%–3.6\% reduction in average accuracy. Scaling analyses further reveal that moderate reverse-group augmentation enhances alignment robustness, whereas excessive augmentation yields diminishing returns due to over-diversification. Overall, DGPO offers a data-efficient approach that balances directional consistency and reasoning diversity across domains.

% This work explores reasoning alignment through directionally structured supervision and introduces \emph{Directional-Groupwise Preference Optimization (DGPO)}, an extension of preference optimization that explicitly models directional consistency while preserving intra-group diversity. 
% Built upon curated forward–reverse exemplars, our framework constructs structured groups that capture both coherent and contrastive reasoning behaviors, enabling more fine-grained alignment beyond single-solution supervision. 
% Empirical results demonstrate that DGPO consistently improves alignment across multiple model families, achieving accuracy gains of 2.2\%–3.6\%. 
% Ablation studies confirm that both uncertainty regularization and direction-aware modeling are beneficial, as removing either leads to a 1.8\%–3.6\% reduction in average accuracy. 
% Scaling analyses further reveal that moderate reverse-group augmentation enhances alignment robustness, whereas excessive augmentation yields diminishing returns due to over-diversification. 
% In summary, DGPO provides a robust and data-efficient framework for reasoning alignment, effectively balancing directional consistency and reasoning diversity across domains.

%% file: sections/6_Limitations.tex
\section{Limitations}

While the proposed DGPO framework benefits from the bidirectional data construction that explicitly introduces forward–reverse reasoning pairs, our dataset generation pipeline is not entirely free from imperfection. Although the solutions are verified for correctness by a model, the constructed problems themselves are not strictly validated for logical or semantic reversibility. Consequently, a portion of the reverse problems may not represent true inverses of their forward counterparts, and some may even lack well-defined answers under the intended reasoning direction.
Despite this limitation, DGPO remains robust in practice: its groupwise contrastive formulation and uncertainty-aware modeling effectively mitigate the influence of such imperfect data, leading to stable optimization and consistent performance gains.

\section{Acknowledgement}
This work is supported by the Advanced Materials-National Science and Technology Major Project (Grant No. 2025ZD0620100), HKUST(GZ)-IEIP-RoP (G01RF000256), and National Key R\&D Program of China (No. 2024YFA1012700).

%% file: sections/7_appendix.tex
\section{Appendix}
\newtcolorbox{promptbox}{
  colback=gray!5,
  colframe=gray!50,
  boxrule=0.5pt,
  arc=2pt,
  left=5pt,
  right=5pt,
  top=5pt,
  bottom=5pt,
  fontupper=\ttfamily\footnotesize,
  enhanced,
  breakable 
}

\subsection{Additional Related Work}
\label{app:related-scaling}
\paragraph{Data Quality Over Scale in Reasoning Alignment.}
A recurring finding in reasoning alignment is that supervision quality and structural diversity often matter more than sheer scale. Less is More for Reasoning (LIMO)~\citep{Ye2025} demonstrates that as few as 817 carefully curated exemplars can yield substantial gains in reasoning performance and generalization. Comparable observations arise in efficiency-oriented scaling~\citep{muennighoff2025s1}, where smaller models achieve performance on par with much larger ones when supported by high-quality exemplars. At the inference level, diversification techniques such as Self-Consistency~\citep{Wang2022sc}, Tree-of-Thoughts~\citep{Yao2023ToT}, Graph-of-Thoughts~\citep{Besta2024GoT}, and Atom-of-Thoughts~\citep{teng2025atom}, together with tool-augmented approaches like ReAct~\citep{Yao2023ReAct}, Toolformer~\citep{Schick2023Toolformer}, and PAL~\citep{Gao2023PAL}, further demonstrate that aggregating diverse reasoning trajectories enhances robustness. On the training side, methods such as self-distillation~\citep{Zelikman2022STAR}, verifier-based filtering~\citep{Cobbe2021TrainingVerifiers}, Self-Play training instances synthesis~\citep{liang2025beyond} and feedback-driven frameworks including Constitutional AI~\citep{Bai2022ConstitutionalAI,ouyang2022training} show that structured supervision can substitute for brute-force scaling. Nevertheless, most existing approaches remain confined to \emph{forward-only} supervision, rely heavily on powerful teacher models, or incur substantial costs for filtering high-quality data.

% \subsection{Theoretical Properties of DGPO}
% \label{app:theory}

% We provide formal statements supporting the discussion in Section~\ref{sec:theory}.

% \begin{proposition}[Consistency]
% If each group $\mathcal{G}^{+}(x)$ and $\mathcal{G}^{-}(x)$ contains one element, 
% Eq.~(\ref{eq:dgpo}) reduces to a pairwise contrastive loss:
% \[
% \mathcal{L}_{\mathrm{DGPO}}
% = -\,\mathbb{E}_{x}\!\left[
% \log \sigma\!\big(\beta (u^{+}(x, y^{+}) - u^{-}(x, y^{-}))\big)
% \right].
% \]
% \end{proposition}

% \begin{proposition}[Smooth Lower Bound]
% The groupwise margin satisfies 
% \[
% A_{\theta}^{+}(x) - A_{\theta}^{-}(x)
% \le
% \max_{y^{+},y^{-}}(u^{+}(x, y^{+}) - u^{-}(x, y^{-})),
% \]
% with equality when the dominant pair determines the group score.
% \end{proposition}

% \begin{proposition}[Expectation Interpretation]
% Let $d \!\sim\! q(d\!\mid\!x,y)$. Then 
% {\small
% \[
% \mathcal{J}(\theta, \phi)
% \approx
% -\,\mathbb{E}_{x,y,d}
% \!\left[
% \log \sigma\!\big(
% \beta (A_{\theta}^{+}(x, d) - A_{\theta}^{-}(x, d))
% \big)
% \right].
% \]
% }

% which shows DGPO maximizes the expected directional alignment likelihood.
% \end{proposition}

\subsection{Implementation in Experiments}
\label{head}

\textbf{Hidden Representation.} For each concatenated sequence $(x;y)$, the hidden state corresponding to the last non-padding token in the final transformer layer is extracted as $h_\theta(x,y)$, identified via the attention mask. A stop-gradient operation is applied to the posterior features in all experiments to prevent gradient flow in subsequent computations.

\textbf{Posterior Head.} A linear projection layer maps the hidden state to two parameters $(\alpha, \beta)$ for each response. A softplus activation ensures the positivity of these parameters, forming a Beta posterior distribution $q(d \mid x, y) = \mathrm{Beta}(\alpha, \beta)$. The mean $d(x,y) = \alpha/(\alpha+\beta)$ and variance $\sigma^2(x,y) = \alpha\beta / [(\alpha+\beta)^2(\alpha+\beta+1)]$ are derived from this distribution.

\textbf{Preference Score.} The preference score is defined as the length-normalized log-probability of the response tokens under the policy model, expressed as $\Delta_\theta(y \mid x) = \frac{1}{|y|} \log \pi_\theta(y \mid x)$. Notably, no reference model is employed in this setup, and thus no log-ratio term is included.

\textbf{Variance-Informed Score Aggregation.} Uncertainty is directly incorporated into the scoring mechanism through the pre-activation terms:
\[
u^{+} = \tau_{\mathrm{win}}^{-1} \Delta_\theta + \log d - c \sigma^2 , 
\]
\[
u^{-} = \tau_{\mathrm{lose}}^{-1} \Delta_\theta + \log(1-d) - c \sigma^2,
\]
with $\tau_{\mathrm{win}} = 0.8$, $\tau_{\mathrm{lose}} = 1.2$, and $c = 1.0$. These per-response scores are aggregated at the group level using a temperature-scaled log-sum-exp operation:
\[
A_\theta^{+} = \tau_{\mathrm{win}} \log \sum_{y \in \mathcal{G}^{+}} e^{u^{+}},\]
\[ \quad A_\theta^{-} = \tau_{\mathrm{lose}} \log \sum_{y \in \mathcal{G}^{-}} e^{u^{-}}.
\]

\textbf{Group Gating.} In the experimental implementation, a gated margin is applied to modulate group-level scores:
\[
g_{\mathrm{win}}(x) = \frac{1}{|\mathcal{G}^{+}|} \sum_{y \in \mathcal{G}^{+}} d(x,y) ,
\]
\[
g_{\mathrm{lose}}(x) = \frac{1}{|\mathcal{G}^{-}|} \sum_{y \in \mathcal{G}^{-}} (1 - d(x,y)).
\]
A gated margin is then constructed as
\[
M_\theta(x) = (A_\theta^{+} - A_\theta^{-}) + \gamma \big[\log g_{\mathrm{win}} - \log g_{\mathrm{lose}}\big],
\]
with $\gamma = 1.0$. 
The contrastive loss is formulated as $-\log \sigma(\lambda_{\mathrm{margin}}\, M_\theta(x))$, 
where $\lambda_{\mathrm{margin}}>0$ is the logistic scaling factor.

\textbf{KL Regularization.} A directional Beta prior is applied to regularize the posterior: $\mathrm{Beta}(2,1)$ for winning responses $(y \in \mathcal{G}^{+})$, encouraging $d \to 1$, and $\mathrm{Beta}(1,2)$ for losing responses $(y \in \mathcal{G}^{-})$, encouraging $d \to 0$. The strength of the KL divergence penalty is controlled by a coefficient $\lambda_{\mathrm{kl}}$.

\textbf{Variance Penalty.} An additional penalty term, weighted by a coefficient $\lambda_{\mathrm{var}}^{(\mathrm{grp})}$, is applied to the average posterior variance $\sigma^2$. When variance is already integrated into the score aggregation, this standalone penalty is reduced by an order of magnitude to prevent excessive regularization. 

\textbf{Beta Prior Parameters for Directional KL Penalty}

The directional KL divergence penalty $\mathcal{R}_{\mathrm{KL}}$ defined in Equation~\ref{eq:rkl} utilizes asymmetric Beta distributions as informative priors. The specific parameter choices are as follows:

\begin{itemize}
    \item \textbf{Preferred group prior ($p_{+}$):} We use $\mathrm{Beta}(2, 1)$ for the preferred group $\mathcal{G}^{+}$. This distribution has a mode at $(2-1)/(2+1-2) = 1$ and mean $2/3 \approx 0.67$, placing strong probability mass near 1 to encourage high consistency estimates.
    
    \item \textbf{Dispreferred group prior ($p_{-}$):} We use $\mathrm{Beta}(1, 2)$ for the dispreferred group $\mathcal{G}^{-}$. This distribution has a mode at $(1-1)/(1+2-2) = 0$ and mean $1/3 \approx 0.33$, concentrating probability density near 0 to favor low consistency estimates.
\end{itemize}

The KL divergence $\mathrm{KL}(q \parallel p)$ measures the information loss when using prior $p$ to approximate the posterior estimate $q$. By minimizing this divergence with the asymmetric priors described above, the estimator is regularized to produce values aligned with our domain knowledge: consistently high for preferred solutions and consistently low for dispreferred ones. This contrasts with a uniform $\mathrm{Beta}(1, 1)$ prior, which provides no directional guidance.

The weighting coefficient $\lambda_{\mathrm{kl}}$ controls the strength of this regularization and is typically set to $1.0$ unless otherwise specified in ablation studies.

\subsection{Prompt Templates and Generation Settings}
\label{app:prompt}
We use the following decoding and sampling parameters throughout data generation:
\begin{verbatim}
"messages": messages,
"think_budget": 8192,
"max_tokens": 2048,
"temperature": 0.6,
"top_p": 0.95,
"top_k": 30,
"stream": False,
\end{verbatim}

The first template is used in the main experiments to generate reverse reasoning problems.
\begin{promptbox}
"role\_definition": \\
"You are an AI model tasked with generating a reflective thinking exercise. \\
Given the following question and answer:" \\

- \textbf{Question}: \{question\} \\
- \textbf{Answer}: \{answer\} \\

"instructions": \\
"Your task is to reverse the roles of the question and answer. \\
Transform the answer into a question that is thought-provoking and encourages deeper reflection. \\
Similarly, convert the original question into a statement that serves as an insightful answer. \\
Ensure that the new question remains reasonable and stimulates further inquiry, \\
while the new answer is right to the question." \\

"expected\_output": \\
- \textbf{New Question}: \\
- \textbf{New Answer}:
\end{promptbox}

The second template elicits step-by-step reasoning and final answers from Qwen3-32B, serving as a basis for high-quality reasoning examples.

\begin{promptbox}
"role\_definition": \\
"You are an AI model that is designed to generate solutions to a given question. \\
All numerical answers must be explicitly marked with \textbackslash boxed\{\}." \\[0.5em]

- \textbf{Question}: \{question\} \\[0.5em]

"instructions": \\
"Ensure your answer is absolutely correct and standard." \\[0.5em]

"expected\_output": \\
Presents the complete and concise answer. \\
If the answer contains only one numerical value, it must be marked in the form of \textbackslash boxed\{\}.
\end{promptbox}

The third template is used to fact-check model-generated answers. It focuses on analytical verification, encouraging explicit reasoning and a concise binary verdict on correctness.

\begin{promptbox}
You are a meticulous fact-checking assistant.\\
1. Carefully reason through the model's answer to the given question.\\
2. Use relevant knowledge, logical reasoning, or explicit calculations to support your analysis.\\
3. After reaching a conclusion, output exactly two clean lines as follows:\\
   - JUDGE: <yes|no> \\
     ('yes' if the model's verdict is factually correct, 'no' otherwise.)\\
Question:\\
\{question\}\\
Model verdict (yes/no):\\
\{model's answer\} 
\end{promptbox}

The fourth template aims to construct reverse reasoning problems derived from verified forward examples.
\begin{promptbox}
You are an expert mathematical problem designer. \\
Given:\\
Original Problem:\\
\{question\}\\
Original Answer:\\
\{model's answer\} \\
Your task:\\
Create 3 reverse problems inspired by this original problem.\\
Each reverse problem must:\\
1. Be fully specified with no hidden or missing conditions.\\
2. Have exactly one unique correct answer, supported by clear reasoning for uniqueness.\\
3. Be meaningfully connected to the original problem by inverting knowns and unknowns, modifying parameters, or extending constraints.\\
Return four problems in the following structured format:\\
Problem 1\\
- Statement:\\
- Answer:\\
Problem 2\\
- Statement:\\
- Answer:\\
Problem 3\\
- Statement:\\
- Answer:
\end{promptbox}

\begin{table*}[ht]
\centering
\caption{
Reverse Data Quality Evaluation.
}
\label{tab:reversedata quality-evaluation}
\resizebox{\textwidth}{!}{
\begin{tabular}{lccccccc}
\toprule
\textbf{Dataset} 
& \textbf{Base Model} 
& \textbf{AIME-25} 
& \textbf{GPQA} 
& \textbf{Math 500} 
& \textbf{GMQ} 
& \textbf{LMGH} 
& \textbf{Avg. Acc.} \\
\midrule
Subset 1 &  Qwen3-1.7B-Base & 3.3\%  & 29.8\% & 46.2\%    & 33.6\%  &  12.9\% &  25.2\% \\
Subset 2 & Qwen3-1.7B-Base &  6.7\% & 29.3\% & 46.6\%  &33.6\%&  10.2\% &  25.3\% \\
Subset 3 &  Qwen3-1.7B-Base & 3.3\% &   28.3\% &  48.2\%  & 35.3\%  & 7.5\%  &  24.5\%\\
\midrule
Subset 1 &  Qwen3-1.7B & 6.7\%  &  27.3\%  &34.8\% &32.5\% &5.2\%  & 21.3\%\\
Subset 2 & Qwen3-1.7B &  3.3\% & 25.3\% & 32.2\%&36.6\% &4.5\% & 20.4\% \\
Subset 3 &  Qwen3-1.7B & 6.7\% &   29.3\% & 40.2\%&31.3\% &5.3\%  & 22.6\%\\
\bottomrule
\end{tabular}
}
\end{table*}

\begin{table*}[h]
\centering
\caption{Statistical experiment information of DGPO averaged over three independent runs.}
\label{tab:multi-run-study}
\resizebox{\textwidth}{!}{
\begin{tabular}{lrrrrr}
\toprule
\textbf{Base Model} 
& \textbf{AIME-25} 
& \textbf{GPQA} 
& \textbf{Math 500} 
& \textbf{GMQ} 
& \textbf{LMGH}  \\
\midrule
Qwen3-1.7B-Base 
& $6.7\% \pm 11.1\%$ 
& $28.6\% \pm 0.7\%$
& $47.5\% \pm 1.0\%$ 
& $35.2\% \pm 0.2\%$ 
& $3.5\% \pm 0.5\%$ \\
Qwen3-1.7B (Official) 
& $25.6\% \pm 4.7\%$ 
& $31.6\% \pm 0.6\%$
& $46.9\% \pm 1.6\%$ 
& $36.3\% \pm 0.6\%$ 
& $13.7\% \pm 2.8\%$ \\
\bottomrule
\end{tabular}
}
\end{table*}
% \begin{table*}[h]
% \centering
% \caption{
% Performance of DGPO on the RL-aligned Qwen3-1.7B model.
% The best results are highlighted in \textbf{bold}.
% }
% \label{tab:appendix-instruct}
% \resizebox{\textwidth}{!}{
% \begin{tabular}{llrrrrrr}
% \toprule
% \textbf{Base Model} & \textbf{Training Strategy}
% & \textbf{AIME-25} 
% & \textbf{GPQA} 
% & \textbf{Math 500} 
% & \textbf{GMQ} 
% & \textbf{LMGH} 
% & \textbf{Avg. Acc.} \\
% \midrule
% Qwen3-1.7B & None   & 13.3\%  & 29.8\%  & \textbf{47.6\%}  & 36.2\%  & 10.6\%  & 27.5\% \\
% Qwen3-1.7B & DGPO (Ours) & \textbf{26.7\%} & \textbf{32.3\%} & 46.0\% & \textbf{36.5\%} & \textbf{12.9\%} & \textbf{30.9\%} \\
% \bottomrule
% \end{tabular}
% }
% \end{table*}
\subsection{Multi-run Robustness of DGPO}
\label{app:multi-run}
\paragraph{Examples of Reverse Problem Construction}
To illustrate how reverse problems are constructed from original forward problems, 
we provide below one representative case.  

\textbf{Original problem:} Find the arithmetic mean of all three-digit palindromes (numbers that read the same forward and backward

\textbf{Original answer:} 550.

\textbf{Reverse problems:}

\begin{enumerate}
    \item 
    \emph{Given that the arithmetic mean of all three-digit palindromes is 550, find their total sum.}  

    \item 
    \emph{There are 90 three-digit palindromes in total. Find the remainder when the largest three-digit palindrome (999) is divided by this number.}  

    \item 
    \emph{How many three-digit palindromes cannot be expressed as the arithmetic mean of two other distinct three-digit palindromes?}  
\end{enumerate}

These examples demonstrate how reverse supervision is systematically constructed: 
each reverse problem maintains a close semantic link to the original forward problem 
while introducing a new perspective (e.g., altering the unknown, parameterizing radii, 
or changing tangency relations).

To assess stability, we repeat DGPO training three times for each base model family (Qwen3-1.7B-Base, SFT(Mixed), and Qwen3-1.7B), reporting mean accuracy and standard deviation across runs. Results are shown in Table~\ref{tab:multi-run-study}. The small variance across runs suggests that DGPO yields consistent gains independent of random initialization and training noise.

\subsection{Implication Details}
\label{app:impl}

We implement all experiments within the \texttt{SWIFT} framework~\citep{zhao2024swiftascalablelightweightinfrastructure}. Supervised fine-tuning (SFT) is performed on the Qwen3-1.7B-Base model using the standard \texttt{SWIFT} framework, trained for 3 epochs with a learning rate of $1\times10^{-5}$. DGPO is implemented by extending the framework with a custom \texttt{DGPO} module that augments the groupwise objective with posterior modulation. Training is performed for 2 epochs with a learning rate of $3\times10^{-6}$ for Qwen3-1.7B-Base and $1\times10^{-6}$ for Qwen3-1.7B, per-device batch size 1, and gradient accumulation 1 (global batch size of 4). All experiments employ DeepSpeed ZeRO-3~\citep{rajbhandari2020zero} in bfloat16 precision for memory efficiency, running on 4$\times$RTX A6000 (48GB) GPUs. For SFT, the maximum sequence length is set to 11,000 tokens to capture the full reasoning traces distilled from the teacher model. For DGPO training, we compare three initialization settings: (1) continuing from the SFT-distilled model, (2) training directly on Qwen3-1.7B-Base, and (3) training directly on Qwen3-1.7B. Depending on the dataset, training takes between one and four hours. In all DGPO cases, the maximum sequence length is capped at 1,000 tokens to balance long-form coverage with the computational feasibility of preference optimization.

\subsection{Reverse Data Quality Details}
\label{app:reverse-quality}

\paragraph{Reverse Data Distillation.}
Table \ref{tab:reversedata quality-evaluation} reports the performance of SFT models trained on three reverse subsets constructed from bidirectional data. Across benchmarks, models distilled on reverse data achieve accuracy levels comparable to those obtained with the forward-only LIMO dataset, indicating that the generated reverse problems maintain a similar level of supervision quality. However, for RL-aligned backbones, performance declines consistently across all benchmarks, suggesting that additional distillation on reverse data may interfere with previously optimized reward-aligned behaviors. These findings indicate that while reverse supervision offers strong groupwise learning signals that benefit base models, its advantages do not seamlessly extend to RL-aligned ones. Incorporating directional objectives into the reinforcement alignment process calls for further exploration to fully harness the potential of group-level reverse data.

\subsection{Qualitative Comparison with DPO Variants}
\label{app:dpo-case-study}

To further illustrate the difference between DGPO and pairwise preference baselines, we include the qualitative case study added during rebuttal. The task asks: \emph{Find the sum of all integer bases $b > 9$ for which $17_b$ is a divisor of $97_b$.}

\paragraph{DGPO answer chain.}
DGPO first converts the base-$b$ expressions into their base-10 forms, then derives the divisibility condition that $b+7$ must divide $9b+7$, which is equivalent to requiring $b+7$ to divide 56. It finally filters the valid divisors by the base constraint $b > 9$, subtracts 7 from each admissible divisor, and sums the resulting bases.

\paragraph{Vanilla DPO answer chain.}
Vanilla DPO identifies the divisibility condition and lists the divisors of 56 correctly, but it fails in the final filtering step. In particular, it retains invalid candidates such as 16, 25, and 36, which do not satisfy the full divisibility requirement, leading to an incorrect final set of bases.

\paragraph{Base-model answer chain.}
The unaligned Qwen3-1.7B-Base model identifies the initial condition and simplifies the expression, but it also keeps invalid bases that violate the constraint $b > 9$, ultimately producing an incorrect total of 275.

This example highlights the qualitative advantage of DGPO: the groupwise directional signal does not merely identify the right algebraic condition, but also helps the model preserve the final consistency check needed to rule out superficially plausible yet invalid solutions.